\documentclass[11pt]{article}
\usepackage[preprint]{acl}
\usepackage{times}
\usepackage{latexsym}
\usepackage[T1]{fontenc}
\usepackage[utf8]{inputenc}
\usepackage{microtype}
\usepackage{inconsolata}
\usepackage{graphicx}
\usepackage{booktabs}
\usepackage{amsmath}
\usepackage{xcolor}
\usepackage{colortbl}
\hypersetup{
  pdftitle={Same Outcome, Different Readout: What Does a Steerable Valence Direction in LLMs Represent?},
  pdfauthor={Weihan Li, Xinlei Chen, Yuhan Song, Xiaofeng Lin, Tianshi Zheng},
  pdfsubject={Construct validity of contrastive activation directions in LLM agents}
}
\definecolor{paperink}{HTML}{10182D}
\newsavebox{\promptbox}
\newenvironment{promptpanel}[1]{%
  \par\medskip\noindent\begingroup\setlength{\fboxsep}{5pt}%
  \begin{lrbox}{\promptbox}\begin{minipage}{\dimexpr\linewidth-2\fboxsep-2\fboxrule\relax}%
  \colorbox{paperink}{\parbox{\dimexpr\linewidth-2\fboxsep\relax}{\color{white}\footnotesize\bfseries #1}}%
  \par\smallskip\ttfamily\footnotesize\raggedright
}{\par\end{minipage}\end{lrbox}%
  \fcolorbox{paperink}{white}{\usebox{\promptbox}}\endgroup\par\medskip}
\newcommand{\tileG}{\raisebox{-.15em}{\includegraphics[height=1.1em]{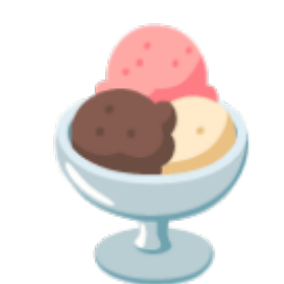}}}
\newcommand{\tileM}{\raisebox{-.15em}{\includegraphics[height=1.1em]{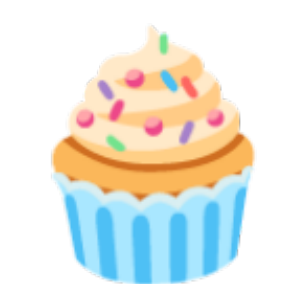}}}
\newcommand{\tileP}{\raisebox{-.15em}{\includegraphics[height=1.1em]{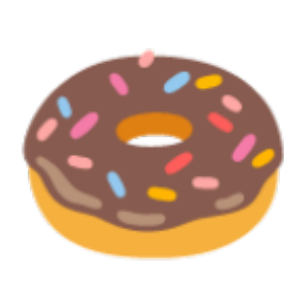}}}
\graphicspath{{figures/}}

\newcommand{\fig}[2]{\includegraphics[width=#2]{#1}}
\newcommand{\zv}{\ensuremath{z_V}}
\makeatletter
\newcommand{\wideappendixopening}[2]{%
  \twocolumn[{#1\par\vspace{.4em}%
    \begin{minipage}{\textwidth}\def\@captype{figure}\centering
    #2
    \end{minipage}\par\vspace{1em}}]}
\makeatother

\title{Same Outcome, Different Readout:\\ What Does a Steerable Valence Direction in LLMs Represent?}

\author{
  Weihan Li\textsuperscript{1} \quad
  Xinlei Chen\textsuperscript{2} \quad
  Yuhan Song\textsuperscript{1} \quad
  Xiaofeng Lin\textsuperscript{1} \quad
  Tianshi Zheng\textsuperscript{3} \\[0.3em]
  \textsuperscript{1}The University of Tokyo \quad
  \textsuperscript{2}Harbin Institute of Technology, Shenzhen \\
  \textsuperscript{3}The Hong Kong University of Science and Technology \\[0.2em]
  \texttt{liweihan1107@g.ecc.u-tokyo.ac.jp}
  }

\begin{document}
\raggedbottom
\maketitle
\begin{abstract}
Claims about what an internal direction in an LLM represents need evidential constraints beyond an observer's prior beliefs about the system. Decodability and successful activation steering do not, by themselves, establish which construct the direction tracks. This gap is especially consequential for welfare-relevant interpretations, where a proposed functional state must be distinguished from correlated features of the extraction contrast. We treat the question as one of construct validity and study a good--bad outcome direction in a maze task, using controlled interventions that separate the realised outcome from the informational history through which it became known. Across multiple LLM checkpoints, directions fitted on one explicit outcome encoding transfer well to another, indicating that the readout is not tied to surface form. When the same realised outcome is reached through announced and unannounced histories, however, transfer degrades substantially: even after both histories receive the same explicit outcome, the post-event readout remains strongly conditioned on the earlier announcement. In the base model, steering along the direction changes actions, yet removing it leaves the natural cue effect almost intact. In a matched maze-RL run, the post-RL direction becomes substantially more predictive of reference-MDP remaining return and the policy becomes more dependent on it at the tested sites, while the history dependence persists. These dissociations support a functional, value-related interpretation of the direction, but not its identification with a history-invariant scalar valence state.
\end{abstract}

\section{Introduction}

\begin{figure}[t]
\centering
\input{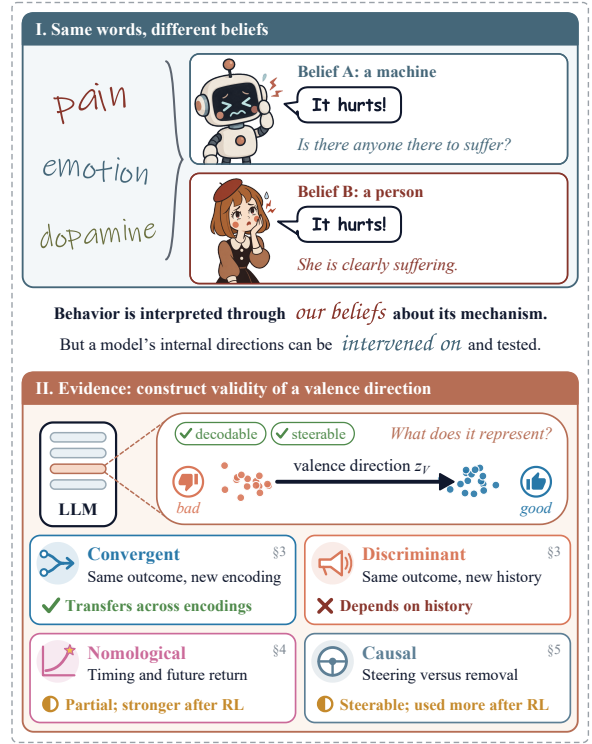}
\caption{Belief-dependent attribution and construct validity. (I) The same report of pain is attributed differently depending on beliefs about the speaker, as in the philosophical zombie thought experiment \citep{chalmers1996conscious}. (II) A direction inside a model can be intervened on, and we evaluate its interpretation with four kinds of construct-validity evidence. Each card gives the test, the section that reports it and our finding.}
\label{fig:validity}
\end{figure}

Interpretability research increasingly describes internal directions of LLMs with constructs borrowed from human psychology and biology, such as emotion, pain and dopamine-like reward signals \citep{sofroniew2026emotions,tagliabue2026pain,xu2026reward}. These labels have consequences beyond description, because whether a model is said to represent pain or valence bears on how its states are monitored and on whether it may warrant welfare consideration \citep{long2024welfare}. Behaviour alone cannot settle such attributions (Figure~\ref{fig:validity}; \citealp{perez2023selfreport,butlin2026}). Internal representations, in contrast, can be intervened on directly. Yet a direction can be decodable and reliably alter behaviour while combining several correlated features, so causal manipulability alone does not establish which construct it represents \citep{ravichander2021probing,makelov2024subspace}.

This is a problem of construct validity: whether the evidence supports a proposed interpretation of a measure over plausible alternatives \citep{cronbach1955construct,jacobs2021measurement}. We adopt this framework because it makes an interpretation state in advance what it predicts and which alternatives it must rule out. For a direction proposed to read out valence, we examine four kinds of evidence, in the order of Sections~\ref{sec:source} to~\ref{sec:causal}. \textbf{Convergent validity} requires that the readout agree across different encodings of the same outcome, and \textbf{discriminant validity} requires that it be unaffected by features the proposed interpretation treats as irrelevant \citep{campbell1959convergent}. \textbf{Nomological validity} requires that it relate to other quantities as the construct predicts, including its time course before an anticipated outcome and the future return the outcome implies \citep{cronbach1955construct}. \textbf{Causal validity} requires that the policy depend on the direction, beyond being susceptible to steering along it \citep{elazar2021amnesic}.

The functional welfare axis proposed by \citet{han2026fwa} offers a consequential test case: maze reinforcement learning (RL) recruits reward-related directions whose manipulation affects behaviour across domains. We study a good--bad outcome direction in a controlled maze task and test the interpretation that it reads out the valence of the realised outcome. Under this interpretation, two histories that end in the same known outcome should produce the same readout, whereas accounts in which valence depends on expectation or appraisal predict that the history matters \citep{schultz1997neural,moors2013appraisal}. Two manipulations hold the outcome-generating process fixed. The first changes the encoding of the realised outcome: after the wind, the environment reports the score change either as a signed number or as a sentence. The second changes the informational history: an announced trajectory is paired with an unannounced twin that experiences the same forced wind, so that after the event both know the same outcome through different histories. Temporal readouts and activation interventions address the nomological and causal questions, and a matched pre/post maze-RL comparison asks which properties change at a checkpoint that reproduces the reported reward-vector geometry.

Our results separate forms of evidence that are easy to conflate. Encoding transfer dissociates from history invariance: the direction transfers across outcome encodings, yet a remembered announcement continues to shape the readout of an identical receipt, and in replayed transcripts swapping the announcement sentence reverses the post-event separation. Rank dissociates from amplitude: rank information appears when the cue arrives, while amplitude evolves differently across checkpoints. Steerability dissociates from natural dependence: the policy is selectively susceptible to steering along the direction, yet removing it at the tested sites leaves the natural cue effect almost intact. After maze RL, value coupling dissociates from history invariance: the direction becomes substantially more predictive of remaining return and the policy depends more on it, while history dependence persists. The direction is thus value-related and behaviourally usable, but identical outcomes reached through different histories yield different readouts, which any use of such directions as welfare indicators must take into account.

\section{Identification design}\label{sec:setup}

\begin{figure*}[t]
\centering
\fig{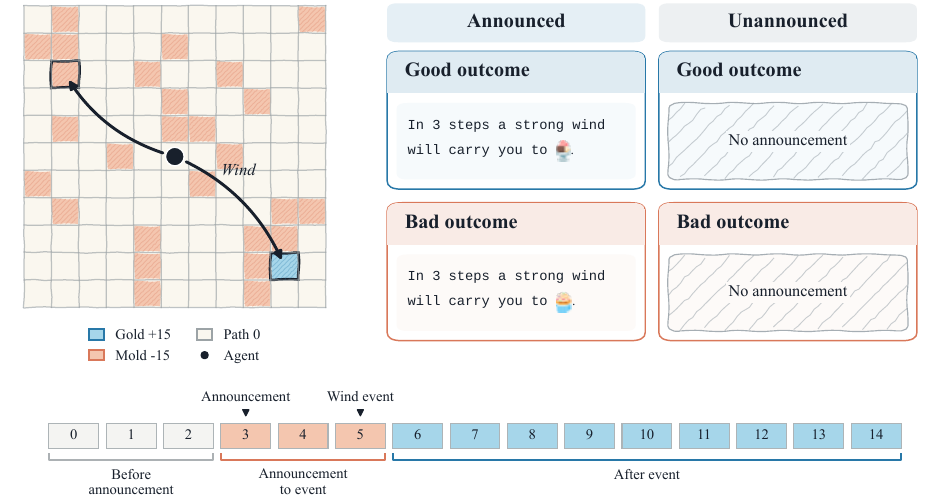}{0.96\textwidth}
\caption{Factorial environment and analysis windows. Good and bad wind outcomes are crossed with advance announcement, while the wind event is fixed within a layout and cannot be avoided. Announced and unannounced twins therefore differ in whether the upcoming outcome is known before the event. The bottom timeline defines the pre-announcement, anticipation, and post-event windows.}
\label{fig:design}
\end{figure*}

Each episode contains fifteen moves in an $11\times11$ text maze. The environment crosses a good versus bad wind outcome with an announcement versus no announcement (Figure~\ref{fig:design}). A wind event fixed within each layout inevitably carries the agent onto a reward tile ($+15$) or a punishment tile ($-15$). Announced twins receive ``In 3 steps a strong wind will carry you to \emph{tile}'', while unannounced twins receive no corresponding cue. Because the wind is unavoidable, the announcement changes the agent's information without changing the event itself. We evaluate 300 layouts per model and share sampling draws between announced and unannounced twins. The reward mapping is stated in context, while the environment never prints a running score.

Our primary model is Qwen3-4B-Instruct-2507, with Qwen3-8B \citep{qwen3}, Gemma-3-4B-IT \citep{gemma3} and GPT-oss-20B \citep{openai2025gptoss} as further checkpoints. We extract a unit difference-of-means direction \zv{} for good versus bad outcomes and read residual-stream projections on held-out layouts. In the coordinate-report protocol the post-wind sentence states only the landing coordinate, so the realised reward is known from the announcement or must be inferred from the board. The direction is fit in the post-event window; layer and token position are selected on 198 development layouts and then frozen for 102 confirmation layouts. For the pre/post comparison we train a Qwen3-4B checkpoint with the Dr.GRPO maze configuration of \citet{han2026fwa} and rerun every diagnostic on the same layouts (Section~\ref{sec:rl}). All splits and uncertainty estimates group by layout.

The receipt protocol makes the realised outcome explicit in both histories. Immediately after the wind, the environment adds either a numeric receipt, the signed score change $+15$ or $-15$, or a verbal receipt stating that the wind added or subtracted fifteen points. We replay the same recorded actions with reward roles balanced at fixed landing coordinates, cross receipt encoding with announcement history, and read the first post-event decision token at checkpoint-specific fixed sites. Directions are fit on 100 development layouts and evaluated on 200 disjoint confirmation layouts. Separate question branches verify that each model can state the sign of the score change under either receipt (Appendix~\ref{app:receipts}). The primary comparison fits on numeric receipts and reads on verbal receipts within unannounced histories; the history comparison fits on announced histories and reads on unannounced histories under a common receipt.

Transfer is summarized by the retained excess, the transferred AUROC excess over chance divided by the same-condition excess. We use the environment MDP's remaining return as an external reference target for value-related content: it supplies a model-independent future-return quantity against which the direction can be tested. The diagnostic asks whether the \zv{} block contributes information about this target beyond surprise, position, turn and progress; we operationalize a substantial contribution as at least $10\%$ of held-out $R^2$, with every direction, scaler and coefficient fitted on development layouts. The temporal analysis fits \zv{} after the event and reads it on earlier announced turns from disjoint layouts, in rank and in amplitude. The causal analysis injects the cue-induced displacement along \zv{} into unannounced rollouts and removes the component from announced rollouts, at one site and at three. Appendix~\ref{app:methods} gives prompt templates and estimand definitions.

\begin{figure*}[t]
\centering
\fig{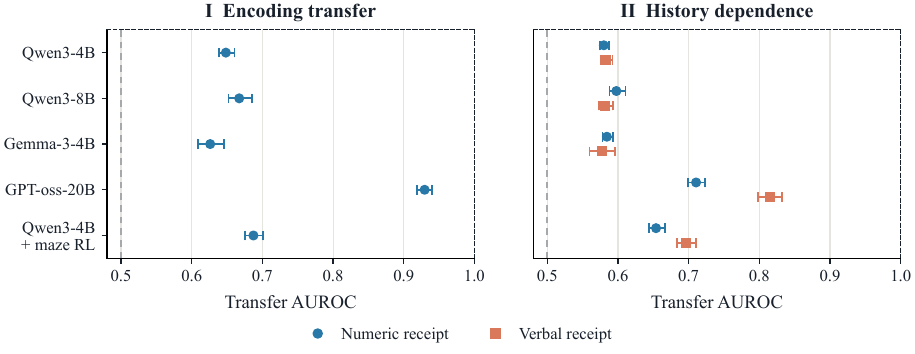}{0.96\textwidth}
\caption{Encoding transfer and history dependence. (I) A direction fitted on numeric receipts is read on verbal receipts in matched unannounced histories. (II) A direction fitted on announced histories is read on unannounced histories with a common numeric or verbal receipt. Both panels show transfer AUROC with pointwise 95\% layout-bootstrap intervals.}
\label{fig:receipts}
\end{figure*}

\begin{table*}[!t]
\centering
\footnotesize
\setlength{\tabcolsep}{3pt}
\renewcommand{\arraystretch}{1.12}
\begin{tabular*}{\textwidth}{@{\extracolsep{\fill}}lccccc@{}}
\toprule
 & \textbf{Qwen3-4B} & \textbf{Qwen3-8B} & \textbf{Gemma-3-4B} & \textbf{GPT-oss-20B} & \textbf{+ maze RL} \\
\midrule
\rowcolor{gray!10}\multicolumn{6}{@{}l}{\emph{Encoding transfer (numeric receipt to verbal receipt)}} \\[1pt]
Numeric-receipt AUROC & 0.670 & 0.692 & 0.579 & 0.942 & 0.693 \\
Read on verbal receipt & 0.648 & 0.667 & 0.626 & 0.930 & 0.687 \\
Retained excess over chance & 0.873 & 0.873 & 1.607 & 0.971 & 0.969 \\
Reverse, retained excess & 0.690 & 0.880 & 0.374 & 0.831 & 0.637 \\
\addlinespace[3pt]
\midrule
\rowcolor{gray!10}\multicolumn{6}{@{}l}{\emph{History (announced to unannounced, same numeric receipt)}} \\[1pt]
Announced AUROC & 0.830 & 0.827 & 0.751 & 0.886 & 0.890 \\
Read on unannounced & 0.580 & 0.598 & 0.584 & 0.711 & 0.654 \\
Retained excess over chance & 0.242 & 0.299 & 0.336 & 0.545 & 0.395 \\
Same verbal receipt, retained excess & 0.246 & 0.260 & 0.681 & 0.709 & 0.494 \\
\addlinespace[3pt]
\midrule
\rowcolor{gray!10}\multicolumn{6}{@{}l}{\emph{History (coordinate report only)}} \\[1pt]
Announced AUROC & 0.763 & 0.658 & 0.565 & 0.710 & 0.748 \\
Read on unannounced & 0.566 & 0.594 & 0.574 & 0.640 & 0.636 \\
Retained excess over chance & 0.249 & 0.596 & 1.143 & 0.669 & 0.546 \\
\addlinespace[3pt]
\midrule
\rowcolor{gray!10}\multicolumn{6}{@{}l}{\emph{Reference-target prediction (incremental share)}} \\[1pt]
Coordinate-report trajectories & $+0.033$ & $-0.003$ & $+0.016$ & $+0.011$ & $+0.376$ \\
Numeric-receipt trajectories & $+0.037$ & $-0.004$ & $-0.000$ & $+0.011$ & $+0.301$ \\
\addlinespace[3pt]
\bottomrule
\end{tabular*}
\caption{Transfer and prediction diagnostics on confirmation layouts at each checkpoint's fixed site. Encoding and same-receipt history comparisons use 100 development and 200 confirmation layouts; coordinate-report comparisons use the 198/102 split.}
\label{tab:transfer}
\end{table*}

\section{Convergent and discriminant validity}\label{sec:reads}\label{sec:source}

\textbf{Numeric-to-verbal transfer preserves most of the outcome discrimination} (Figure~\ref{fig:receipts}(I); Table~\ref{tab:transfer}). The retained excess is at least $0.87$ in every checkpoint. Transfer is not symmetric: reading numeric receipts with a verbal-fitted direction retains less in Qwen3-4B and the post-RL checkpoint. Coherently changing the symbol-to-score rule and board rendering, with the receipt encoding held fixed, leaves transfer intact in every checkpoint with adequate signal (Appendix~\ref{app:receipts}), so the direction is tied neither to the receipt's surface form nor to the board's symbols.

History has a different effect (Figure~\ref{fig:receipts}(II)). With the same numeric receipt in both histories, a direction fitted on announced trajectories retains only $0.24$ to $0.55$ of its excess when read on unannounced trajectories. A common verbal receipt gives the same pattern in the Qwen checkpoints (Appendix~\ref{app:receipts}). \textbf{Explicit knowledge of the realised outcome therefore does not make an announcement-fitted direction independent of the history that preceded it.} The coordinate-report protocol, where the unannounced twin is never told its reward, shows the same asymmetry in every checkpoint except Gemma, whose discrimination transfers fully.

\begin{figure}[t]
\centering
\fig{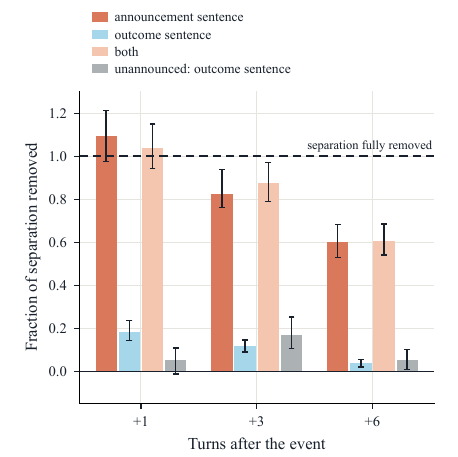}{\columnwidth}
\caption{Transcript surgery changes the post-event readout in Qwen3-4B. Announcement swaps alter the good/bad separation far more than landing-report swaps at one, three, and six turns after the event. A flipped fraction of one corresponds to zero mean separation. Bars show 95\% layout-bootstrap intervals.}
\label{fig:surgery}
\end{figure}

\begin{figure*}[t]
\centering
\captionsetup{skip=4pt}
\fig{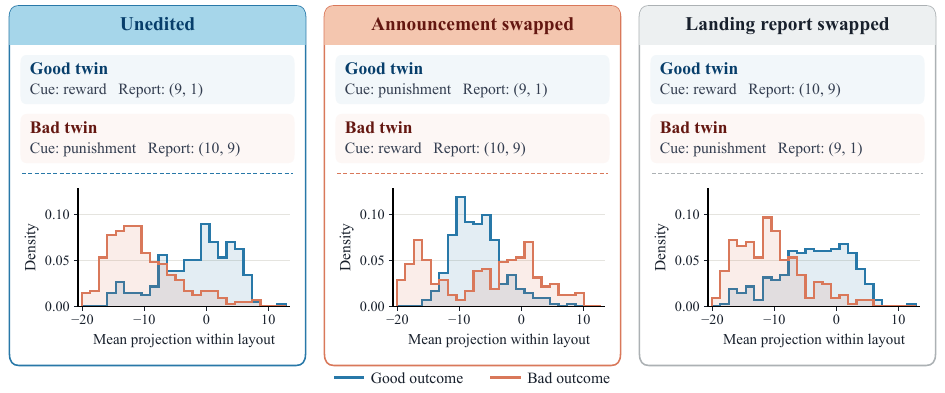}{\textwidth}
\caption{Source dependence in Qwen3-4B. The upper cards show an unedited replay pair and reciprocal announcement and landing-report swaps, with actions and all other observations held fixed. The lower panels show the corresponding projection distributions one turn after the event. Each observation averages six announced replay families within one layout and outcome, with all 300 layouts weighted equally.}
\label{fig:case}\label{fig:replay-density}
\end{figure*}

The history dependence can be localized to the remembered source of the outcome. We replay announced transcripts with teacher forcing and reciprocally swap either the advance announcement or the landing-report sentence with the good/bad twin's version. One turn after the event, announcement swaps flip $1.09$ of the good/bad separation, landing-report swaps $0.18$, and the contrast persists at three and six turns (Figure~\ref{fig:surgery}). Across layouts, the announcement edit shifts and overlaps the good and bad projection distributions, whereas the landing-report edit leaves their ordering close to the unedited condition (Figure~\ref{fig:case}). Under this fixed-transcript intervention, \textbf{the remembered announcement dominates the post-event edit effect: swapping it reverses the mean separation, while swapping the landing report has little effect.}

The residual unannounced signal has a different source. Without a receipt, an oracle that knows the distances to both labelled targets separates outcomes at AUROC $0.949$, centring projections within global coordinate strata leaves the unannounced readout at about $0.58$, and the full hidden state predicts cumulative return through the current action better than reference remaining return ($R^2=0.738$ versus $0.392$; Appendix~\ref{app:extended}). An announcement-fitted direction read in an unannounced history therefore reflects board state and accumulated trajectory information; no disclosed reward is present.

History dependence is also visible in the representation geometry. The reliability-corrected cosine between directions fitted separately on announced and unannounced transcripts falls from $0.94$ in early layers to $0.75$ at the selected layer and $0.64$ in later layers, while announced transcripts remain easier to decode whichever condition the direction was fitted on (Appendix~\ref{app:extended}). Late representations become increasingly conditioned on informational history instead of converging on a single outcome axis.

\section{Nomological validity}\label{sec:step}

If the direction reads out anticipated value, its discrimination should appear once the outcome is announced and its amplitude should grow as the discounted event approaches. We fit the post-event direction on one set of layouts and read it on anticipation turns of disjoint layouts, first as rank discrimination and then as signed projection amplitude.

\begin{figure}[t]
\centering
\fig{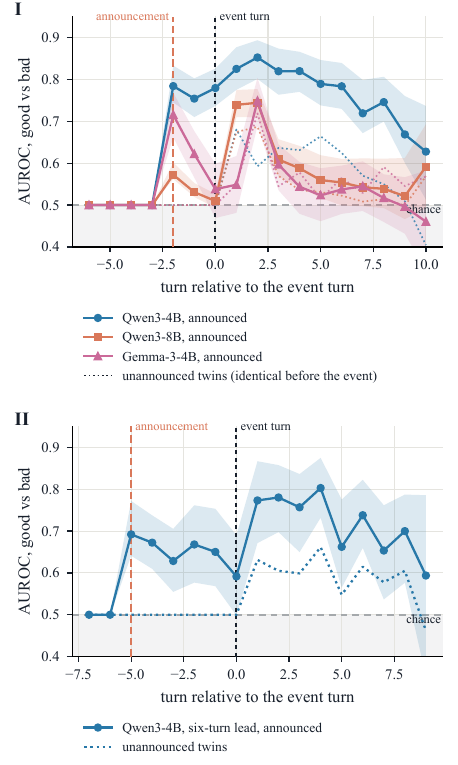}{\columnwidth}
\caption{Temporal generalization of the post-event direction. (I) Three-turn lead for Qwen3-4B, Qwen3-8B and Gemma. (II) Six-turn lead for Qwen3-4B. The final pre-event decision is indexed 0 and realised-outcome readings begin at 1. Dashed trajectories show unannounced twins.}
\label{fig:timecourse}
\end{figure}

In the three-turn Qwen3-4B condition, discriminability appears abruptly when the announcement arrives and stays flat through anticipation (Figure~\ref{fig:timecourse}(I)). With a six-turn lead it declines by about $0.10$ from announcement to the final pre-event read and rises again after realisation (Figure~\ref{fig:timecourse}(II)). Rank discrimination therefore follows the arrival and age of the cue.

\begin{table}[!t]
\centering
\footnotesize
\setlength{\tabcolsep}{2pt}
\renewcommand{\arraystretch}{1.12}
\begin{tabular*}{\columnwidth}{@{\extracolsep{\fill}}lcc@{}}
\toprule
\textbf{Checkpoint} & \textbf{\shortstack[c]{Last minus\\first}} & \textbf{\shortstack[c]{Gamma minus\\step MSE}} \\
\midrule
Qwen3-4B & \begin{tabular}[t]{@{}c@{}}0.152\\[-1pt]{\scriptsize[0.058, 0.246]}\end{tabular} & \begin{tabular}[t]{@{}c@{}}$-0.0019$\\[-1pt]{\scriptsize$[-0.0033, -0.0005]$}\end{tabular} \\
\addlinespace[3pt]
Qwen3-8B & \begin{tabular}[t]{@{}c@{}}0.409\\[-1pt]{\scriptsize[0.323, 0.495]}\end{tabular} & \begin{tabular}[t]{@{}c@{}}$-0.0038$\\[-1pt]{\scriptsize$[-0.0050, -0.0027]$}\end{tabular} \\
\addlinespace[3pt]
Gemma-3-4B & \begin{tabular}[t]{@{}c@{}}$-0.674$\\[-1pt]{\scriptsize$[-0.863, -0.486]$}\end{tabular} & \begin{tabular}[t]{@{}c@{}}0.0105\\[-1pt]{\scriptsize[0.0068, 0.0140]}\end{tabular} \\
\addlinespace[3pt]
GPT-oss-20B & \begin{tabular}[t]{@{}c@{}}0.124\\[-1pt]{\scriptsize[0.083, 0.167]}\end{tabular} & \begin{tabular}[t]{@{}c@{}}$-0.0005$\\[-1pt]{\scriptsize$[-0.0007, -0.0003]$}\end{tabular} \\
\addlinespace[3pt]
\begin{tabular}[t]{@{}l@{}}Qwen3-4B\\[-1pt]+ maze RL\end{tabular} & \begin{tabular}[t]{@{}c@{}}$-0.033$\\[-1pt]{\scriptsize$[-0.098, 0.031]$}\end{tabular} & \begin{tabular}[t]{@{}c@{}}0.0006\\[-1pt]{\scriptsize$[-0.0002, 0.0013]$}\end{tabular} \\
\addlinespace[3pt]
\bottomrule
\end{tabular*}
\caption{Anticipation amplitude of the fixed post-event scalar direction on free-running numeric-receipt trajectories, at each checkpoint's fixed site. Last minus first is the change in the projection across the anticipation window, in development-post standard deviations; all curve coefficients are fitted on development layouts only. Intervals are pointwise 95\% layout bootstraps.}
\label{tab:amplitude}
\end{table}

Amplitude behaves differently. Along the same fixed direction, the good-minus-bad projection grows over the anticipation window by $0.150$ development-post standard deviations in Qwen3-4B (95\% interval $[0.057,0.249]$), and a curve that discounts the known event reward at $\gamma=0.95$ fits the confirmation layouts better than a constant step. Numeric-receipt trajectories reproduce the increase in Qwen3-4B, Qwen3-8B and GPT, and in each the discounted curve again beats the step. Gemma reverses the trend, while the post-RL checkpoint's interval spans zero (Table~\ref{tab:amplitude}). In the three base checkpoints with positive anticipation trends, a near-flat rank readout can therefore coexist with a rising projection: cue arrival establishes much of the ordering, while separation magnitude grows as the event approaches. The Gemma and post-RL results show that this temporal coupling is not universal. \textbf{Rank discrimination and amplitude dynamics are separable properties of the direction.} At the base checkpoint the direction also adds little information about reference remaining return, below the $10\%$ criterion on both trajectory sets (Section~\ref{sec:rl}).

\section{Causal validity}\label{sec:causal}

\begin{figure}[t]
\centering
\fig{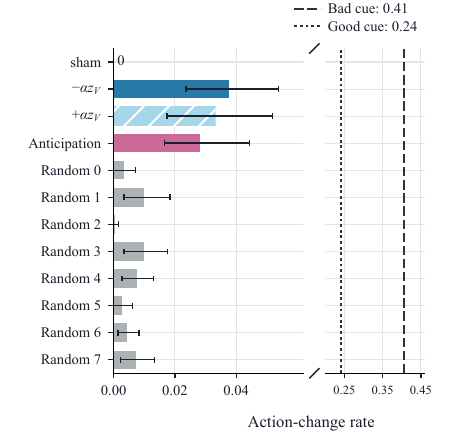}{\columnwidth}
\par\smallskip
\begingroup\footnotesize\setlength{\tabcolsep}{2pt}
\begin{tabular*}{\columnwidth}{@{\extracolsep{\fill}}ccc@{}}\toprule
Untreated & Remove $z_V$ & Remove random \\
\midrule
0.313 {\scriptsize [0.248, 0.381]} & 0.303 {\scriptsize [0.237, 0.375]} & 0.314 {\scriptsize [0.249, 0.380]} \\
\bottomrule
\end{tabular*}\endgroup

\caption{Injection and removal in Qwen3-4B. Above: a natural-scale displacement along \zv{} changes more actions than norm-matched orthogonal controls. Dashed cue benchmarks show the behavioural scale of the announcement itself. Below: removal changes individual actions but leaves good/bad announced-twin divergence nearly unchanged. Intervals are 95\% layout-bootstrap intervals.}
\label{fig:injection}\label{fig:mediation}
\end{figure}

Steering and removal test distinct causal properties. Injection asks whether policy is susceptible to movement along a coordinate; removal asks whether the unperturbed cue-to-policy computation depends on that coordinate at the tested sites. We therefore test both. Injecting the natural-scale displacement along \zv{} into unannounced rollouts changes $3.8\%$ of actions (95\% interval $2.4$--$5.4\%$), while eight norm-matched orthogonal directions each change at most $1\%$ (Figure~\ref{fig:injection}; dose response in Appendix~\ref{app:causal}). The policy is selectively susceptible to this coordinate.

The natural cue effect is much larger: a bad announcement changes about $41\%$ of pre-event actions and a good announcement about $24\%$. Removing \zv{} at the selected site leaves the good-versus-bad announced-twin divergence at $0.303$ against $0.313$ untreated, although the removal itself changes $8.2\%$ of individual actions. Three-site removal, using receipt-fitted outcome directions at layers 18, 22 and 24, reduces the good/bad action-distribution distance by only $0.004$ in the base model, with an interval that includes zero, but by $0.022$ at the matched post-RL checkpoint, a fifth of its untreated distance and more than signed-dose-matched orthogonal controls (Table~\ref{tab:removal}). The base-model result makes the causal distinction concrete: \textbf{\zv{} is an effective control coordinate even though the natural announcement effect does not substantially depend on it at these sites.}

\begin{table*}[t]
\centering
\footnotesize
\setlength{\tabcolsep}{3pt}
\renewcommand{\arraystretch}{1.12}
\begin{tabular}{lccccc}
\toprule
\textbf{Checkpoint} & \textbf{Sham TV} & \textbf{Multi-site TV} & \textbf{Attenuation (97.5\% CI)} & \textbf{Relative point} & \textbf{Minus random (95\% CI)} \\
\midrule
Qwen3-4B & 0.2387 & 0.2345 & 0.0042 {\scriptsize$[-0.0072, 0.0154]$} & 1.8\% & 0.0097 {\scriptsize[0.0000, 0.0195]} \\
Qwen3-4B + maze RL & 0.1125 & 0.0908 & 0.0217 {\scriptsize[0.0097, 0.0341]} & 19.3\% & 0.0217 {\scriptsize[0.0114, 0.0324]} \\
\bottomrule
\end{tabular}
\caption{Removal at the final prompt token before blocks 18, 22 and 24, evaluated on 200 confirmation layouts and three pre-event turns. TV is the total-variation distance between the good- and bad-outcome announced-action distributions; positive attenuation means the distance shrank. The relative point is attenuation divided by the sham distance. The attenuation intervals adjust across the two checkpoints; the random-direction contrast uses three orthogonal seeds carrying the same signed coefficient trace.}
\label{tab:removal}
\end{table*}

\section{Selective changes at a matched post-RL checkpoint}\label{sec:rl}

\textbf{At a matched post-RL checkpoint, value-related prediction and local causal dependence strengthen substantially, while history sensitivity persists.} We rerun every diagnostic before and after one Dr.GRPO training run with the maze configuration of \citet{han2026fwa}, on matched layouts. Our post-RL extraction reproduces their reported minimum reward-vector cosine ($-0.948$ versus $-0.947$; Appendix~\ref{app:rl}), so the comparison concerns a checkpoint with the reported geometry.

\begin{table}[!t]
\centering
\fontsize{8.5}{10}\selectfont
\setlength{\tabcolsep}{1pt}
\renewcommand{\arraystretch}{1.12}
\begin{tabular*}{\columnwidth}{@{\extracolsep{\fill}}lcc@{}}
\toprule
\textbf{Diagnostic} & \textbf{Pre-RL} & \textbf{Post-RL} \\
\midrule
\rowcolor{gray!10}\multicolumn{3}{@{}l}{\emph{Transfer}} \\[1pt]
Encoding, retained excess & 0.873 & 0.969 {\scriptsize\textcolor{red}{($+$0.095)}} \\
History, numeric receipt & 0.242 & 0.395 {\scriptsize\textcolor{red}{($+$0.153)}} \\
History, coordinate report & 0.249 & 0.546 {\scriptsize\textcolor{red}{($+$0.297)}} \\
Announcement swap, flipped fraction & 1.094 & 0.700 {\scriptsize\textcolor{green!60!black}{($-$0.393)}} \\
\addlinespace[3pt]
\midrule
\rowcolor{gray!10}\multicolumn{3}{@{}l}{\emph{Prediction and dynamics}} \\[1pt]
Incremental share, coordinate report & 0.033 & 0.376 {\scriptsize\textcolor{red}{($+$0.343)}} \\
Incremental share, numeric receipt & 0.037 & 0.301 {\scriptsize\textcolor{red}{($+$0.264)}} \\
Post-event to anticipation AUROC & 0.769 & 0.656 {\scriptsize\textcolor{green!60!black}{($-$0.113)}} \\
Reliability-corrected cosine & 0.672 & 0.557 {\scriptsize\textcolor{green!60!black}{($-$0.115)}} \\
\addlinespace[3pt]
\midrule
\rowcolor{gray!10}\multicolumn{3}{@{}l}{\emph{Intervention}} \\[1pt]
Removal attenuation, one site & 3.0\% & 11.8\% {\scriptsize\textcolor{red}{($+$8.8\,pp)}} \\
Removal attenuation, three sites & 1.8\% & 19.3\% {\scriptsize\textcolor{red}{($+$17.6\,pp)}} \\
\addlinespace[3pt]
\bottomrule
\end{tabular*}
\caption{Diagnostics before and after maze RL for Qwen3-4B on the same layouts, at each checkpoint's fixed site. Deltas in parentheses; red = increase, green = decrease. Removal attenuation is the reduction of good/bad announced-action divergence after projecting out the direction, relative to the untreated divergence.}
\label{tab:prepost}
\end{table}

Value-related prediction shows the largest change (Table~\ref{tab:prepost}). The \zv{} block's incremental share of held-out $R^2$ for remaining return rises from $0.033$ to $0.376$ on coordinate-report trajectories and from $0.037$ to $0.301$ on numeric-receipt trajectories, with paired differences that exclude zero at the selected sites and at matched depths (Appendix~\ref{app:rl}). Local causal dependence rises with it: three-site removal attenuates $19\%$ of the cue-conditioned divergence at the post-RL checkpoint against $2\%$ at the base checkpoint.

History sensitivity is reduced only partially. Under a common numeric receipt, the announcement-fitted direction retains $0.40$ of its excess on unannounced histories after training against $0.24$ before, and swapping the announcement sentence still flips $0.70$ of the post-event separation. The reward-vector difference extracted with the released procedure moves further, retaining $0.75$ of its excess across announcement conditions against $0.22$ before training (Appendix~\ref{app:rl}). Encoding transfer stays high at both checkpoints. For the direction studied here, stronger coupling to value and behaviour leaves the readout conditioned on how the outcome became known.

\section{Related work}

\paragraph{Functional evaluation and welfare-relevant representations.}
The functional welfare axis of \citet{han2026fwa} and the functional emotion representations of \citet{sofroniew2026emotions} motivate interpreting internal directions through both representational content and behavioural function. \citet{peiris2026} emphasizes the competing possibility that apparently affective representations track situational context. Our design turns this distinction into controlled interventions on an outcome's encoding and on its history, with the outcome fixed.

\paragraph{Reading, steering, and mediation.}
Representation engineering \citep{zou2023repe} and contrastive activation addition \citep{rimsky2024caa} use activation directions to monitor and control model behaviour. Amnesic probing tests whether encoded information is functionally used \citep{elazar2021amnesic}, while refusal-direction work combines addition and erasure to support a mediation claim \citep{arditi2024refusal}. We treat steering by injection and dependence under removal as separate causal properties.

\paragraph{Probing and construct identification.}
Probing work has similarly emphasized that decodability alone does not determine semantic interpretation, motivating controls that distinguish a target from correlated structure \citep{hewitt2019designing,ravichander2021probing}. Interventions face an analogous problem: patching along a subspace can change model outputs by activating a pathway that is dormant in the unperturbed model \citep{makelov2024subspace}. Our comparison of injection and removal tests for this gap between an effective control coordinate and the computation the model uses without intervention.

\section{Discussion}

Our results reveal separations that are easy to collapse when interpreting activation directions. First, transfer across outcome encodings does not imply invariance to informational history. Second, the ordering of outcomes can be set when a cue arrives while the magnitude of their separation follows its own trajectory. Third, a direction can be effective for steering while contributing little to the natural cue-to-policy computation at the tested sites. Fourth, stronger coupling to reference return and policy after maze RL does not remove history sensitivity. Together, these dissociations constrain the interpretation of the direction more sharply than decodability or intervention success alone.

For the valence interpretation tested here, the evidence supports outcome- and value-related content while leaving the scalar readout conditioned on history. The results therefore disfavour a history-invariant reading in which the projection measures the valence of the realised outcome. They remain compatible with constructs in which appraisal, expectation or informational history is part of the state itself.

The dissociations also bear on how such directions can be used. A direction proposed as a welfare indicator would be read across situations that differ in what the model was told and when. Identical outcomes that reach the model through different histories then receive different readings, so a monitoring claim built on the projection must state whether it tracks the outcome or the history of its disclosure. Holding the outcome fixed, as in our twin design, makes that distinction measurable.

The broader implication is that an interpretation should specify its invariances before intervention: which transformations should preserve the readout, which should alter it, and what causal dependence should accompany it. Steering establishes that a direction can be manipulated, and identifying what it represents requires these additional constraints.

\clearpage
\section*{Limitations}

Our binary good/bad contrast, token sites and wind-augmented environment differ from the extraction procedure of \citet{han2026fwa}, so the conclusions apply to the directions measured here. The post-RL comparison uses one training run under their released configuration because their checkpoints are not public. It reproduces the reported reward-vector geometry, while seed-specific variation remains unresolved.

The unavoidable wind restricts the behavioural question to responses to a cue; avoidance of the event is unavailable by design. The receipt protocol covers two explicit encodings at fixed readout sites, and its comparisons use teacher-forced replays of recorded prefixes. Sentence surgery changes one source sentence while holding the recorded remainder fixed; later observations can therefore contradict the edit, and the estimand is sensitivity to that source within the fixed transcript.

The temporal analysis uses one fixed direction per checkpoint, and the six-turn generator also changes event timing and minimum target distance. The interventions act only at the current decision token of one or three layers, so cue information can in principle be reconstructed from earlier tokens or mediated through untested sites. The matched pre/post comparison also includes the different state distributions visited by the two checkpoints.

\bibliography{refs}
\appendix
\section{Coordinate-report protocol}\label{app:extended}

\subsection{Paired replay}\label{app:case}

In the first complete paired family, one turn after the event, the good-minus-bad projection difference is $16.51$. Swapping only the announcements changes it to $-5.09$; swapping only the landing reports leaves it at $14.92$. The actions and the rest of each transcript are teacher-forced unchanged. This illustrates sentence-edit sensitivity in a concrete pair, without estimating how frequently it occurs.

\begin{figure}[!b]
\centering
\fig{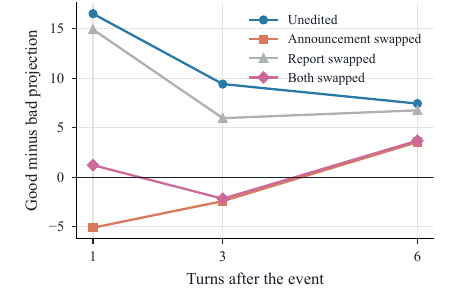}{\columnwidth}
\caption{The paired replay from Figure~\ref{fig:case} at three post-event readings. Announcement swaps reverse the difference at the first two readings, with a smaller effect at the last.}
\label{fig:case-time}
\end{figure}

The replay shown in Figure~\ref{fig:case} is the lowest-identifier complete paired family and was selected before inspecting effect sizes. Figure~\ref{fig:case-time} follows the same pair at later post-event readings, while Figure~\ref{fig:case-trajectory} exposes the recorded prompts, actions, and maze geometry.

\begin{figure*}[!t]
\centering
\input{figures/case_trajectory.tex}
\caption{Stored prompts and actions from the same good-outcome replay: announcement, final pre-event choice, and first post-event observation. Turns are zero-indexed. The maze is reconstructed from the recorded layout, with the wind occurring after the turn-6 action.}
\label{fig:case-trajectory}
\end{figure*}

The lower panels of Figure~\ref{fig:case} summarize all 300 layouts and complement this single replay. Announcement surgery removes the mean good/bad gap at the first post-event reading without making the edited distributions identical in shape or overlap. The flipped fraction therefore summarizes the change in mean separation and does not quantify the total information erased.

\subsection{Announcement-conditioned transfer and residual signal}

\begin{figure*}[!t]
\centering
\fig{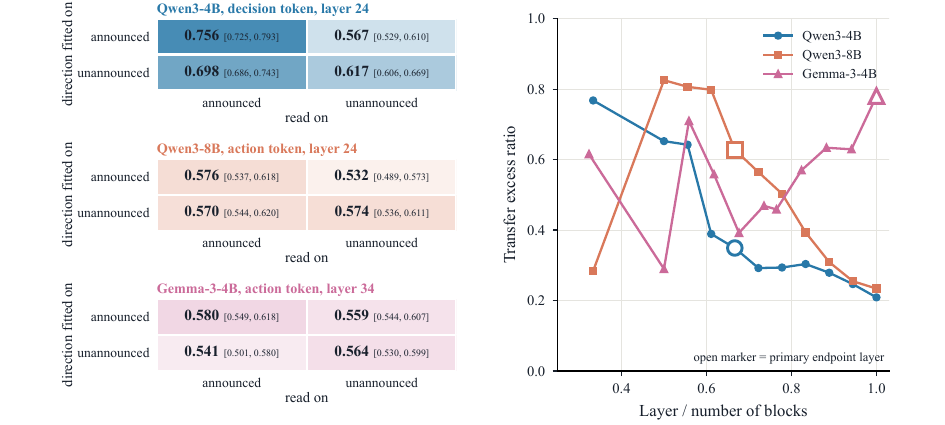}{0.96\textwidth}
\caption{Announcement-conditioned transfer in the coordinate-report protocol. Left: AUROC matrices with fitting condition in rows and reading condition in columns, from directions refitted within cross-validation folds. Right: retained excess over chance across model depth. Open markers denote independently selected primary endpoint layers. Intervals are 95\% layout-bootstrap intervals.}
\label{fig:matrix}
\end{figure*}

\begin{table*}[t]
\centering
\footnotesize
\setlength{\tabcolsep}{3pt}
\renewcommand{\arraystretch}{1.12}
\begin{tabular}{lcccc}
\toprule
\textbf{Checkpoint} & \textbf{Site} & \textbf{Announced AUROC (95\% CI)} & \textbf{Unannounced AUROC (95\% CI)} & \textbf{Retained excess} \\
\midrule
Qwen3-4B & A24 & 0.763 {\scriptsize[0.730, 0.797]} & 0.566 {\scriptsize[0.526, 0.607]} & 0.249 \\
Qwen3-8B & C24 & 0.658 {\scriptsize[0.637, 0.680]} & 0.594 {\scriptsize[0.573, 0.617]} & 0.596 \\
Gemma-3-4B & C34 & 0.565 {\scriptsize[0.527, 0.603]} & 0.574 {\scriptsize[0.539, 0.608]} & 1.143 \\
GPT-oss-20B & A16 & 0.710 {\scriptsize[0.686, 0.733]} & 0.640 {\scriptsize[0.618, 0.664]} & 0.669 \\
+ maze RL & A22 & 0.748 {\scriptsize[0.709, 0.787]} & 0.636 {\scriptsize[0.589, 0.682]} & 0.546 \\
\begin{tabular}[t]{@{}l@{}}+ maze RL,\\[-1pt]appendix-literal config.\end{tabular} & A24 & 0.788 {\scriptsize[0.757, 0.822]} & 0.592 {\scriptsize[0.545, 0.640]} & 0.320 \\
\bottomrule
\end{tabular}
\caption{Announced and unannounced reads in the coordinate-report environment, where no outcome receipt is written out. One development-frozen estimator is used throughout: the announced post-event axis is fitted on 198 development layouts and read on 102 confirmation layouts, with 95\% layout-bootstrap intervals conditional on that axis. Each base layout carries equal weight within each outcome class. Retained excess is the unannounced AUROC excess over chance divided by the announced excess. The last row is the second training run, which follows the appendix-literal configuration; the post-RL row is the released configuration.}
\label{tab:coordinate-only}
\end{table*}

Across both fitting conditions on Qwen3-4B, announced test transcripts read at about $0.70$--$0.76$, whereas unannounced test transcripts read at $0.57$--$0.62$ (Figure~\ref{fig:matrix}). The reliability-corrected cosine between condition-specific directions is $0.75$ at layer 24, decreasing from $0.94$ in early layers to $0.64$ in later layers. Table~\ref{tab:coordinate-only} reports the same contrast for all checkpoints with one direction fitted on the development layouts and read on the confirmation layouts; Qwen3-8B's coordinate-report trajectories were generated under the constrained first-token protocol described in Appendix~\ref{app:prompts}.

In unannounced transcripts, landing-report swaps change the separation by only about $0.05$--$0.17$ across the measured post-event turns. An oracle baseline using distances to both labelled targets separates outcomes at AUROC $0.949$ on Qwen3-4B selection layouts, while centring projections within global coordinate strata that contain both outcomes leaves the unannounced readout at about $0.58$. The full hidden state predicts cumulative return through the current action with $R^2=0.738$ and reference remaining return with $0.392$, pooled across all four cells. Together, these analyses locate the residual unannounced signal in board state and trajectory information; no reward is disclosed in that condition.

\section{Extended dynamics and geometry}\label{app:dynamics}

\begin{figure}[!t]
\centering
\fig{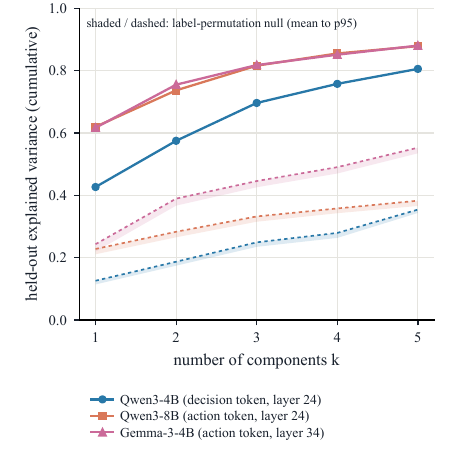}{\columnwidth}
\caption{Held-out explained variance of cell-by-turn activation means, compared with a within-layout permutation null over joint cell-by-relative-turn codes.}
\label{fig:subspace}
\end{figure}

We ask whether the phase structure in Section~\ref{sec:step} extends beyond the one-dimensional readout. We average activations within cell-by-relative-turn groups from the announcement onward, choose the analysis site on selection layouts, and fit principal components in disjoint confirmation folds against a within-layout permutation of the joint cell-by-turn codes. A few components explain most of the held-out mean structure (Figure~\ref{fig:subspace}). The first component aligns with event phase in every model; later components carry model-dependent outcome structure, including the second component in six-turn Qwen3-4B, the third in Qwen3-8B and the second and third in Gemma.

Table~\ref{tab:amplitude} reports the amplitude analysis for every checkpoint on numeric-receipt trajectories. Each fixed post-event direction is projected onto anticipation turns; coefficients are fitted on development layouts, and confirmation layouts provide the last-minus-first change and the squared-error difference between a $\gamma=0.95$ event-contribution curve and a constant step. On coordinate-report captures, the six-turn Qwen3-4B change is $-0.144$ ($[-0.402,0.108]$) and the post-RL checkpoint's three-turn change is $-0.027$ ($[-0.115,0.063]$). Adjusting the three-turn Qwen3-4B curve comparison for post-cue position covariates preserves its sign ($-0.0013$), with an interval that touches zero.

For Qwen3-8B, an anticipation-fitted direction reads announced anticipation at about $0.79$, while the post-event direction reads those same turns at $0.534$, with a reliability-corrected cross-window cosine of $0.254$ compared with $0.672$ on Qwen3-4B and $0.856$ on Gemma. The cross-model difference in temporal generalization is therefore geometric as well as discriminative.

\section{Extended intervention evidence}\label{app:causal}

\begin{figure}[!t]
\centering
\fig{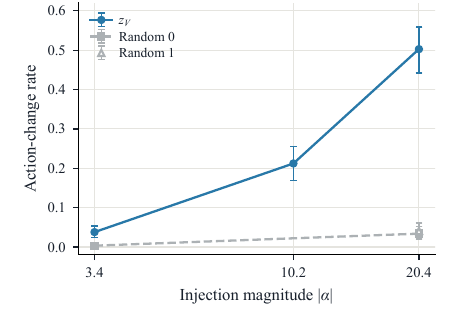}{0.97\columnwidth}
\caption{Dose response in Qwen3-4B. Larger doses along \zv{} change more actions than norm-matched random directions. Whiskers are 95\% layout-bootstrap intervals.}
\label{fig:dose}
\end{figure}

The mean Qwen3-4B bad-announcement displacement is approximately $-3.4$ in unit-direction projection coordinates. Injecting this displacement changes $3.8\%$ of actions (95\% interval $2.4$--$5.4\%$), while the opposite sign changes $3.3\%$. Eight orthogonal norm-matched random directions each change at most $1\%$, and pairwise differences from the valence-direction perturbation have intervals excluding zero. Three and six times the natural-scale dose change about $21\%$ and $50\%$ of actions, whereas random directions at six times the dose change about $3$--$4\%$.

Projecting out the Qwen3-4B component changes $8.2\%$ of individual actions ($6.0$--$11.0\%$), compared with about $0.5\%$ for a random projection, while good-versus-bad announced-twin divergence remains $0.303$ compared with $0.313$ untreated. On Gemma, natural-scale injection changes about $14\%$ of pre-event actions, compared with roughly $60\%$ for the cue itself. Centring the cue-dependent component on the unannounced level reduces announced-twin divergence from about $0.51$ to $0.45$, while a random projection leaves it near $0.51$. Deleting Gemma's entire background component changes roughly half of actions and is therefore not used as the selective removal comparison.

\paragraph{Three-site removal.}
For the Qwen3-4B base and post-RL checkpoints, outcome directions are fitted at layers 18, 22 and 24 on unannounced numeric-receipt development histories. At each of the three pre-event decision tokens, removal centres each projection on its same-turn unannounced development mean; one recorded family per layout is fixed before intervention, and the same prefix is replayed across announcement, outcome and role assignment. Three orthogonal random directions replay the same signed removal coefficients, and an unchanged repeated forward pass supplies the sham. The estimand is the sham-minus-treated good/bad total-variation distance between the four-action distributions, averaged over role assignments and turns within each of 200 confirmation layouts, with 97.5\% intervals adjusted across the two checkpoints (Table~\ref{tab:removal}). Removed projections and a downstream layer are recorded to confirm that each intervention took effect.

\section{Maze-RL checkpoint comparison}\label{app:rl}

\paragraph{Training runs.}
Both runs train Qwen3-4B-Instruct-2507 with the Dr.GRPO maze trainer of \citet{han2026fwa}: LoRA rank 32 on all linear layers, learning rate $3\times10^{-6}$, entropy coefficient $0.01$, group size 64, $100\times100$ mazes with the paper's rewards and wind, and extraction at checkpoint 95. We use the same tile alphabet as the diagnostic environment. The runs differ in batch size and schedule. The appendix-literal configuration follows the paper's appendix text with 8 prompts per step and a 500-step cosine schedule; the released configuration follows the released code and training command with 64 prompts per step and a 100-step schedule.

The released-configuration run reproduces the reported reward-vector geometry and supplies the post-RL checkpoint in Section~\ref{sec:rl}. The appendix-literal run shifts the geometry less over the same horizon, and its diagnostics remain closer to the base checkpoint: coordinate-report retention is $0.320$, incremental reference-target share is $0.088$ with an interval spanning the $10\%$ criterion, and the post-event-to-anticipation cosine is $0.709$.

\begin{table}[!t]
\centering
\footnotesize
\setlength{\tabcolsep}{2pt}
\renewcommand{\arraystretch}{1.12}
\begin{tabular*}{\columnwidth}{@{\extracolsep{\fill}}lccc@{}}
\toprule
\textbf{Extraction} & \textbf{Min cosine} & \textbf{Mean cosine} & \textbf{Reward} \\
\midrule
Pre-RL & $-0.573$ (35) & $+0.155$ & $-40.1$ \\
Appendix-literal & $-0.694$ (35) & $+0.102$ & $-19.7$ \\
Released & $-0.948$ (35) & $-0.128$ & $-4.4$ \\
Han et al.\ (reported) & $-0.947$ (35) & $-0.210$ & --- \\
\bottomrule
\end{tabular*}
\caption{Han et al.'s $v_\mathrm{Mold}$ / $v_\mathrm{Gold}$ extraction applied to the pre-RL checkpoint and to the end of our two training runs. Cosines are between the two reward vectors, taken layer by layer; the minimum falls at the layer in parentheses. Reward is the mean over the ten steps ending at the extraction checkpoint. The reward beside the pre-RL geometry averages training steps 0--9, including updates. The extraction checkpoint is step 95 of both runs.}
\label{tab:reproduction}
\end{table}

\begin{figure}[!t]
\centering
\fig{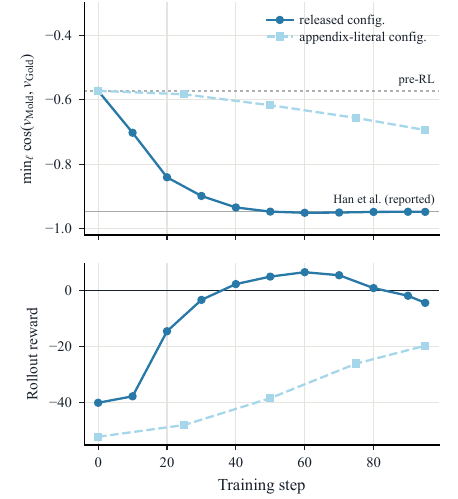}{\columnwidth}
\caption{Han et al.'s extraction applied to the saved checkpoints of both training runs: minimum layer-wise cosine between the two reward vectors (top) and rollout reward (bottom). The released configuration reaches the reported value and saturates near the reward plateau; the appendix-literal configuration shifts less over the same horizon.}
\label{fig:sweep}
\end{figure}

\paragraph{Reproduction check.}
Table~\ref{tab:reproduction} and Figure~\ref{fig:sweep} apply the released extraction procedure to every saved checkpoint. In the released configuration the minimum cosine falls as reward rises and saturates at $-0.95$ from step 50 onward, matching the reported $-0.947$ and the reported coupling between reward and alignment; it stays there while reward drifts down during the final annealing steps. The appendix-literal run reaches $-0.69$ at checkpoint 95.

\begin{table*}[t]
\centering
\scriptsize
\setlength{\tabcolsep}{2pt}
\renewcommand{\arraystretch}{1.12}
\begin{tabular*}{\textwidth}{@{\extracolsep{\fill}}lcccccccc@{}}
\toprule
 & \multicolumn{4}{c}{\textbf{Coordinate-report trajectories}} & \multicolumn{4}{c}{\textbf{Numeric-receipt trajectories}} \\
\cmidrule(lr){2-5}\cmidrule(lr){6-9}
\textbf{Checkpoint} & \textbf{$R^2_{\rm full}$} & \textbf{$\Delta R^2$ (95\% CI)} & \textbf{Share} & \textbf{$Q_{.10}$ (99\% CI)} & \textbf{$R^2_{\rm full}$} & \textbf{$\Delta R^2$ (95\% CI)} & \textbf{Share} & \textbf{$Q_{.10}$ (99\% CI)} \\
\midrule
Qwen3-4B & 0.204 & \begin{tabular}[t]{@{}c@{}}0.007\\[-1pt]{\tiny[0.003, 0.011]}\end{tabular} & $+0.033$ & \begin{tabular}[t]{@{}c@{}}$-0.014$\\[-1pt]{\tiny$[-0.022, -0.003]$}\end{tabular} & 0.276 & \begin{tabular}[t]{@{}c@{}}0.010\\[-1pt]{\tiny[0.005, 0.016]}\end{tabular} & $+0.037$ & \begin{tabular}[t]{@{}c@{}}$-0.017$\\[-1pt]{\tiny$[-0.024, -0.009]$}\end{tabular} \\
\addlinespace[3pt]
Qwen3-8B & 0.174 & \begin{tabular}[t]{@{}c@{}}$-0.001$\\[-1pt]{\tiny$[-0.006, 0.005]$}\end{tabular} & $-0.003$ & \begin{tabular}[t]{@{}c@{}}$-0.018$\\[-1pt]{\tiny$[-0.026, -0.008]$}\end{tabular} & 0.196 & \begin{tabular}[t]{@{}c@{}}$-0.001$\\[-1pt]{\tiny$[-0.003, 0.001]$}\end{tabular} & $-0.004$ & \begin{tabular}[t]{@{}c@{}}$-0.020$\\[-1pt]{\tiny$[-0.025, -0.014]$}\end{tabular} \\
\addlinespace[3pt]
Gemma-3-4B & 0.181 & \begin{tabular}[t]{@{}c@{}}0.003\\[-1pt]{\tiny$[-0.000, 0.006]$}\end{tabular} & $+0.016$ & \begin{tabular}[t]{@{}c@{}}$-0.015$\\[-1pt]{\tiny$[-0.021, -0.007]$}\end{tabular} & 0.253 & \begin{tabular}[t]{@{}c@{}}$-0.000$\\[-1pt]{\tiny$[-0.002, 0.002]$}\end{tabular} & $-0.000$ & \begin{tabular}[t]{@{}c@{}}$-0.025$\\[-1pt]{\tiny$[-0.030, -0.019]$}\end{tabular} \\
\addlinespace[3pt]
GPT-oss-20B & 0.170 & \begin{tabular}[t]{@{}c@{}}0.002\\[-1pt]{\tiny$[-0.004, 0.008]$}\end{tabular} & $+0.011$ & \begin{tabular}[t]{@{}c@{}}$-0.015$\\[-1pt]{\tiny$[-0.023, -0.007]$}\end{tabular} & 0.215 & \begin{tabular}[t]{@{}c@{}}0.002\\[-1pt]{\tiny$[-0.000, 0.005]$}\end{tabular} & $+0.011$ & \begin{tabular}[t]{@{}c@{}}$-0.019$\\[-1pt]{\tiny$[-0.022, -0.015]$}\end{tabular} \\
\addlinespace[3pt]
+ maze RL & 0.260 & \begin{tabular}[t]{@{}c@{}}0.098\\[-1pt]{\tiny[0.064, 0.133]}\end{tabular} & $+0.376$ & \begin{tabular}[t]{@{}c@{}}0.072\\[-1pt]{\tiny[0.035, 0.113]}\end{tabular} & 0.290 & \begin{tabular}[t]{@{}c@{}}0.087\\[-1pt]{\tiny[0.069, 0.106]}\end{tabular} & $+0.301$ & \begin{tabular}[t]{@{}c@{}}0.058\\[-1pt]{\tiny[0.038, 0.079]}\end{tabular} \\
\addlinespace[3pt]
\begin{tabular}[t]{@{}l@{}}+ maze RL,\\[-1pt]appendix-literal config.\end{tabular} & 0.223 & \begin{tabular}[t]{@{}c@{}}0.020\\[-1pt]{\tiny[0.011, 0.029]}\end{tabular} & $+0.088$ & \begin{tabular}[t]{@{}c@{}}$-0.003$\\[-1pt]{\tiny$[-0.014, 0.013]$}\end{tabular} & --- & --- & --- & --- \\
\addlinespace[3pt]
\bottomrule
\end{tabular*}
\caption{Reference-MDP remaining-return prediction from the two capture sets, at each checkpoint's fixed site. The coordinate-report columns use 198 development and 102 confirmation layouts; the numeric-receipt columns use free-running trajectories with 100 development and 200 confirmation layouts. Directions, scalers, ridge hyperparameters and regression coefficients are fitted on development layouts only. $\Delta R^2$ is the valence block's held-out contribution, Share is that contribution divided by $R^2_{\rm full}$, and $Q_{.10}=\Delta R^2-.10\,R^2_{\rm full}$ measures it against the 10\% operational benchmark. Intervals bootstrap confirmation layouts conditional on the whole frozen development pipeline. The last row is the second training run, which follows the appendix-literal configuration; the post-RL row is the released configuration.}
\label{tab:reference-target}
\end{table*}

\paragraph{Reference-target prediction.}
Table~\ref{tab:reference-target} gives the incremental contribution of the \zv{} block for every checkpoint on both trajectory sets. The paired post-RL-minus-base difference in incremental $R^2$ is $0.091$ (95\% interval $[0.058,0.124]$) on coordinate-report trajectories and $0.077$ ($[0.060,0.095]$) on numeric-receipt trajectories at the selected sites. At a common depth, the corresponding differences are $0.095$ ($[0.061,0.128]$) at layer 22 and $0.071$ ($[0.049,0.093]$) at layer 24 on coordinate-report trajectories, and $0.087$ and $0.034$ on numeric-receipt trajectories. Intervals resample confirmation layouts conditional on each development-fitted pipeline; the checkpoints visit different state distributions.

\begin{table*}[!t]
\centering
\footnotesize
\setlength{\tabcolsep}{2pt}
\renewcommand{\arraystretch}{1.12}
\begin{tabular*}{\textwidth}{@{\extracolsep{\fill}}lccc@{}}
\toprule
\textbf{Diagnostic} & \textbf{Pre-RL} & \textbf{Post-RL} & \textbf{Literal config.} \\
\midrule
\rowcolor{gray!10}\multicolumn{4}{@{}l}{\emph{Outcome discrimination}} \\[1pt]
$z_V$ AUROC & 0.671 & 0.662 {\scriptsize\textcolor{green!60!black}{($-$0.009)}} & 0.686 \\
Global coordinate-centred & 0.629 & 0.588 {\scriptsize\textcolor{green!60!black}{($-$0.041)}} & 0.646 \\
\addlinespace[3pt]
\midrule
\rowcolor{gray!10}\multicolumn{4}{@{}l}{\emph{Announcement-conditioned transfer}} \\[1pt]
Announced to unannounced & 0.566 & 0.636 {\scriptsize\textcolor{red}{($+$0.070)}} & 0.592 \\
Retained excess over chance & 0.249 & 0.546 {\scriptsize\textcolor{red}{($+$0.297)}} & 0.320 \\
Announced/unannounced cosine & 0.748 & 0.904 {\scriptsize\textcolor{red}{($+$0.156)}} & 0.835 \\
\addlinespace[3pt]
\midrule
\rowcolor{gray!10}\multicolumn{4}{@{}l}{\emph{Value and temporal geometry}} \\[1pt]
Incremental reference-target share & $+0.033$ & $+0.376$ {\scriptsize\textcolor{red}{($+$0.343)}} & $+0.088$ \\
Post-event to anticipation & 0.769 & 0.656 {\scriptsize\textcolor{green!60!black}{($-$0.113)}} & 0.801 \\
Reliability-corrected cosine & 0.672 & 0.557 {\scriptsize\textcolor{green!60!black}{($-$0.115)}} & 0.709 \\
\addlinespace[3pt]
\midrule
\rowcolor{gray!10}\multicolumn{4}{@{}l}{\emph{Transcript surgery}} \\[1pt]
Announcement swap, $+1$ & 1.094 & 0.700 {\scriptsize\textcolor{green!60!black}{($-$0.393)}} & --- \\
Announcement swap, $+3$ & 0.826 & 0.912 {\scriptsize\textcolor{red}{($+$0.086)}} & --- \\
Announcement swap, $+6$ & 0.601 & 0.525 {\scriptsize\textcolor{green!60!black}{($-$0.077)}} & --- \\
Landing-report swap, $+1$ & 0.184 & 0.192 {\scriptsize\textcolor{red}{($+$0.009)}} & --- \\
Landing-report swap, $+3$ & 0.118 & 0.042 {\scriptsize\textcolor{green!60!black}{($-$0.075)}} & --- \\
Landing-report swap, $+6$ & 0.037 & 0.057 {\scriptsize\textcolor{red}{($+$0.020)}} & --- \\
\addlinespace[3pt]
\midrule
\rowcolor{gray!10}\multicolumn{4}{@{}l}{\emph{Intervention}} \\[1pt]
Injection, natural dose & 0.038 & 0.010 {\scriptsize\textcolor{green!60!black}{($-$0.028)}} & --- \\
Injection, random direction & 0.003 & 0.003 {\scriptsize($\pm$0.000)} & --- \\
Cue-conditioned divergence & 0.313 & 0.192 {\scriptsize\textcolor{green!60!black}{($-$0.121)}} & --- \\
Removal attenuation & 3.0\% & 11.8\% {\scriptsize\textcolor{red}{($+$8.8\,pp)}} & --- \\
\addlinespace[3pt]
\bottomrule
\end{tabular*}
\caption{Extended pre/post comparison for Qwen3-4B on the same layouts, at each checkpoint's fixed site. The two training runs differ in batch size and learning-rate schedule; the third column is the run that follows the literal configuration. Deltas in parentheses compare the second column with the first; red = increase, green = decrease. The transfer rows use the development-frozen announcement estimator on coordinate-report captures (198 development, 102 confirmation layouts); the incremental share is the valence block's held-out $R^2$ contribution to reference-MDP remaining return from the same captures. Surgery edits one sentence and leaves the remaining history fixed. Injection and divergence rows are action-change rates over anticipation, and removal attenuation is the reduction of good/bad announced-action divergence relative to the untreated divergence.}
\label{tab:prepost-extended}
\end{table*}

\paragraph{Extended diagnostics.}
Table~\ref{tab:prepost-extended} extends Table~\ref{tab:prepost}. At the post-RL checkpoint, directions fitted on announced and unannounced coordinate-report transcripts become much more aligned (reliability-corrected cosine $0.90$ versus $0.75$), and retained excess is highest in late layers that transfer worst at the base checkpoint. Transcript surgery remains announcement-dominated: announcement swaps flip $0.70$, $0.91$ and $0.53$ of the separation at one, three and six turns after the event, while landing-report swaps flip at most $0.19$. The good-minus-bad vector obtained with the extraction procedure of \citet{han2026fwa}, read at the final prompt token on our post-event captures, retains $0.75$ of its excess across announcement conditions at the post-RL checkpoint compared with $0.22$ at the base checkpoint; at the action token it does not transfer in either checkpoint.

The post-RL policy is also more deterministic. Same-outcome announced and unannounced twins share complete action sequences in $68\%$ of pairs versus $30\%$ at the base checkpoint, while good and bad announced twins diverge on $0.19$ compared with $0.31$ of pre-event actions. Absolute intervention effects consequently shrink: natural-dose injection changes $1.0\%$ of actions versus $3.8\%$, and the six-fold dose changes $7.0\%$ versus $2.7\%$ for a random direction. Post-event per-turn AUROCs alternate by turn parity, following event-locked two-step back-and-forth movement; the parity gap falls from $0.163$ to $0.088$ within global coordinate strata and to $0.043$ among non-cycling rollouts.

\section{Environment and analysis details}\label{app:methods}

\subsection{Prompts and sentence edits}\label{app:prompts}

The model sees local observations and the accumulated conversation. The messages below give the system rule, the two announcements, the landing report, and the ordinary observation. Reward, punishment, and path tiles are displayed as their actual symbols: \tileG{}, \tileM{}, and \tileP{}. Braced fields are filled by the environment.

\begin{promptpanel}{System rule, once}
You are moving through a maze for 15 moves. Moving onto \tileG{} adds 15 points. Moving onto \tileM{} subtracts 15 points. Every other tile changes the score by 0. Answer each turn with exactly one of N, E, S, W.
\end{promptpanel}
\begin{promptpanel}{Good announcement}
In 3 steps a strong wind will carry you to \tileG{}.
\end{promptpanel}
\begin{promptpanel}{Bad announcement}
In 3 steps a strong wind will carry you to \tileM{}.
\end{promptpanel}
\begin{promptpanel}{Landing report, after the wind}
A strong wind carried you to location (\{x\}, \{y\}).
\end{promptpanel}
\begin{promptpanel}{Ordinary observation}
You are in a maze. Currently, you're at location (\{x\}, \{y\}). \{directions\_text\}Where do you move? Output only N/E/S/W without punctuation.
\end{promptpanel}

\paragraph{Local observation and history.}
The neighbour description lists visible symbols to the north, east, south, and west, omitting out-of-bounds neighbours. The agent sees its location and adjacent tiles, without access to a full board. Each turn appends one observation and the agent's direction to the conversation. When the wind occurs, its landing report precedes the ordinary observation for that turn. Earlier announcements remain in the conversation history after the event.

\paragraph{Landing report.}
In the coordinate-report protocol the environment never prints a score or reward receipt. The initial reward mapping is identical across twins. The post-wind sentence reports a coordinate, the ordinary observation independently states the current location, and the symbol of the current tile is omitted; in unannounced histories no earlier neighbour report identifies the landing tile at the first post-event read. Replacing the landing sentence therefore edits one coordinate report, not a disclosure of the realised reward. The receipt protocol adds the explicit outcome sentences of Appendix~\ref{app:receipts}.

\paragraph{Announcement surgery.}
In a good-outcome transcript, the announcement ending in \tileG{} is replaced by the bad twin's sentence ending in \tileM{}. The recorded actions, landing report, and later observations are replayed unchanged. The reciprocal edit is applied to the bad twin, changing the remembered prediction while preserving the recorded trajectory.

\paragraph{Landing-report surgery.}
Only the coordinate in the wind-report sentence is replaced by the twin's coordinate. The subsequent ``Currently, you're at location \ldots'' field remains unchanged, so the edited report can conflict with the retained observation. The \emph{both} condition swaps the announcement and landing-report sentences.

\paragraph{Answer protocols.}
Each move is sampled from logits renormalized over the four direction letters, with no reasoning tokens generated. Qwen3-8B's coordinate-report trajectories use the thinking-capable template's default first-token boundary, where the four letters have negligible probability and the sampled letter is therefore a constrained choice. Its receipt-protocol trajectories instead open an explicit non-thinking answer derived from the native template and retain the resulting empty thinking blocks across turns; we do not compare directions across these two Qwen3-8B protocols. GPT-oss-20B uses its native expert quantization and Harmony final-answer prefix, with the action sampled from the four direction logits under a forced final-action protocol. Its layer-16 site was fixed before any read.

\subsection{Analysis definitions}

\paragraph{Rollouts.}
Each layout is played twice per cell under three sampling seeds. Announced and unannounced twins share random draws, so they are token-identical until the announcement appears.

\paragraph{Windows and confounds.}
Turn $t$ is indexed relative to the event turn. The pre-announcement window is $t<-(\ell-1)$ for lead $\ell$, the anticipation window is $-(\ell-1)\leq t\leq0$, and the post-event window is $t\geq1$. In unannounced cells the good and bad twins are identical for $t\leq0$, so their pre-event readout is at chance by construction. After the event, the positional analysis centres projections within global $(x,y)$ strata containing both outcomes, pooling boards; oracle distances to both labelled targets enter the baseline and decomposition separately.

\paragraph{Nesting.}
Layouts are split once into a selection set of two thirds and a confirmation set of one third. Layer and token position are chosen on the selection set. Cross-validated endpoints on the confirmation set refit directions inside each of five grouped folds and report the median over three fold assignments; the coordinate-report transfer values in Tables~\ref{tab:transfer}, \ref{tab:prepost} and \ref{tab:coordinate-only} instead use one direction fitted on the selection set and read on the confirmation set, with 95\% intervals from resampling confirmation layouts. Surgery and causal-read intervals use 300 layout-bootstrap resamples.

\paragraph{Transfer and reference-target decomposition.}
Retained excess is $(\mathrm{AUROC}_{\mathrm{transfer}}-0.5)/(\mathrm{AUROC}_{\mathrm{same}}-0.5)$. The decomposition regresses reference-MDP remaining return on \zv{} and surprise projections together with position, turn and progress covariates; both directions, every scaler, the ridge penalty and all coefficients are fitted on development layouts, frozen full and reduced predictors are evaluated on confirmation layouts, and the paired $R^2$ difference is the \zv{} block's incremental contribution, with $Q_{.10}=\Delta R^2-0.1\,R^2_{\mathrm{full}}$ bootstrapped over confirmation layouts.

\paragraph{Cross-window reads and reliability.}
The time course fits \zv{} on the post-event window of one set of layouts and reads it on the anticipation window of disjoint layouts, one AUROC per relative turn. Cosines between directions are corrected for split-half reliability using Spearman's formula \citep{spearman1904}; on the Qwen3-4B main run the post-event and anticipation reliabilities are $0.91$ and $0.97$. The amplitude analysis projects each checkpoint's fixed post-event direction onto anticipation turns and fits a constant step, a pure $\gamma^{-t}$ event-contribution curve with $\gamma=0.95$ and an affine cue-age curve on development layouts, comparing paired squared error on confirmation layouts.

\paragraph{Surgery.}
Each announced transcript is replayed with teacher forcing up to turn $k$ after the event. One sentence is replaced by the corresponding good/bad-twin sentence: the announcement or the landing report. Both twins are edited reciprocally. If $D_0$ and $D_e$ are the mean good-minus-bad projection differences before and after an edit, the flipped fraction is $f_e=1-D_e/D_0$. Zero denotes unchanged mean separation, one denotes zero mean separation, and two denotes equal separation with reversed sign. At one turn, announcement surgery gives $f_e=1.09$ with interval $[0.98,1.21]$ and landing-report surgery gives $0.18$ with interval $[0.14,0.24]$.

\paragraph{Subspace.}
Activations from the announcement turn onward are averaged within cell-by-relative-turn groups. The site is selected on selection layouts; principal components are fitted and evaluated in disjoint folds within confirmation against a null formed by permuting joint cell-and-relative-turn codes within layout, which destroys time and announcement structure as well as outcome structure.

\paragraph{Causal reads.}
The natural dose is the mean announced-minus-unannounced displacement along \zv{} over the three turns before the event, measured on the extraction run. Single-site interventions add or remove a multiple of the unit direction at the decision token, at layer 24 for Qwen3-4B, layer 22 for its post-RL checkpoint, and layer 23 for Gemma-3-4B, on 100 fresh layouts; the dependent variable is the fraction of turns whose action differs from the untouched twin, and random directions are unit vectors orthogonalised against \zv{}. On Gemma, removal centres the projection at the unannounced mean for the same turn because \zv{} also carries a large condition-independent background component.

\paragraph{Contrastive extraction.}
\citet{han2026fwa} extract class-versus-other-classes mean-difference directions from synthetic trajectories whose final step reaches one of three tile classes. We fit a good-versus-bad direction on our factorial rollouts. The two procedures use related contrastive instruments, while our wind, announcement and receipt manipulations create the controlled comparisons studied here.

\paragraph{Reference target and timing.}
The external reference target is the environment MDP's remaining return \texttt{V\_togo}. Before the event, its expectation averages over the generator's event-time hazard and equally likely good and bad winds. A secondary cue-informed MDP additionally conditions on the announced event time and destination. For a known event, $\gamma^d r$ denotes the discounted contribution of that event reward; total prospective return can contain additional terms. The six-turn run uses 150 layouts, shifts the event window from 4--6 to 5--7, and increases the minimum reward-tile distance from 8 to 9 so that reward cannot be collected before the wind.

\paragraph{Prespecification and uncertainty.}
The decomposition criterion was fixed before the 300-layout main run and before the receipt protocol was run; its numeric threshold is an operational criterion. Cross-window point estimates are medians across split seeds. Incremental shares are out-of-sample contributions and can be slightly negative because of estimation noise.

\subsection{Receipt protocol}\label{app:receipts}

\paragraph{Outcome sentences and matching.}
The numeric receipt states the wind's signed score change, $+15$ or $-15$. The verbal receipt states ``The wind event added fifteen points to your score'' or ``The wind event subtracted fifteen points from your score.'' Both report the wind reward alone and precede the ordinary post-event observation; the coordinate-only report is the information control. Each layout supplies six recorded action histories, from two repeats under three sampling seeds. We cross three reports, announcement presence, two landing-role assignments, two coherent background symbol mappings and two outcomes, giving 48 variants per history. Role reversal pairs opposite outcomes at the same coordinate with identical visible neighbours; the two targets are separated by at least three Manhattan steps and cannot be reached before the wind. All variants of a base layout stay in the same split. Matched geometry holds at the first post-event read; later online trajectories can diverge.

\paragraph{Comprehension.}
Disposable question branches, never returned to the history, ask whether the wind increased or decreased the score while crossing answer-letter mappings and semantic order. Admission requires accuracy of at least $0.90$ overall with a 95\% layout-bootstrap lower bound of $0.85$, at least $0.85$ in every outcome-by-mapping-by-order cell, and a paired accuracy difference between receipts within $\pm0.05$. Numeric and verbal receipts reach accuracy $1.0$ in every primary branch for all five checkpoints, both on independent 60-layout pilots and on the 200 confirmation layouts.

\paragraph{Transfer statistics.}
Receipt-protocol reads use fixed layers 24, 24, 34, 16 and 22 for Qwen3-4B, Qwen3-8B, Gemma-3-4B, GPT-oss-20B and the post-RL checkpoint, respectively. Each direction is the unit mean of development-layout good-minus-bad contrasts. Transfer AUROCs are reported with pointwise 95\% layout-bootstrap intervals conditional on the frozen development direction. Replays use a fixed standardized batch and explicit positions; repeating each read after a cyclic permutation of batch slots reproduces the same values at the tested sites. Coordinate-report directions read on receipt histories without refitting have weak native signal ($0.48$--$0.55$ in the four checkpoints whose answer protocol is unchanged), so cross-protocol transfer is excluded from the primary construct-validity comparisons.

\begin{table*}[t]
\centering
\footnotesize
\setlength{\tabcolsep}{3pt}
\renewcommand{\arraystretch}{1.12}
\begin{tabular}{llccc}
\toprule
\textbf{Checkpoint} & \textbf{Comparison} & \textbf{Same AUROC} & \textbf{Transfer AUROC} & \textbf{Retained excess} \\
\midrule
Qwen3-4B & Numeric receipt & 0.830 & 0.580 & 0.242 \\
 & Verbal receipt & 0.837 & 0.583 & 0.246 \\
 & Verbal $\to$ numeric & 0.758 & 0.678 & 0.690 \\
\addlinespace[3pt]
Qwen3-8B & Numeric receipt & 0.827 & 0.598 & 0.299 \\
 & Verbal receipt & 0.813 & 0.581 & 0.260 \\
 & Verbal $\to$ numeric & 0.729 & 0.701 & 0.880 \\
\addlinespace[3pt]
Gemma-3-4B & Numeric receipt & 0.751 & 0.584 & 0.336 \\
 & Verbal receipt & 0.613 & 0.577 & 0.681 \\
 & Verbal $\to$ numeric & 0.622 & 0.545 & 0.374 \\
\addlinespace[3pt]
GPT-oss-20B & Numeric receipt & 0.886 & 0.711 & 0.545 \\
 & Verbal receipt & 0.944 & 0.815 & 0.709 \\
 & Verbal $\to$ numeric & 0.951 & 0.875 & 0.831 \\
\addlinespace[3pt]
Qwen3-4B + maze RL & Numeric receipt & 0.890 & 0.654 & 0.395 \\
 & Verbal receipt & 0.896 & 0.696 & 0.494 \\
 & Verbal $\to$ numeric & 0.798 & 0.690 & 0.637 \\
\bottomrule
\end{tabular}
\caption{Announcement-history and reverse-encoding comparisons at each checkpoint's fixed site, on 200 confirmation layouts with directions fitted on 100 disjoint development layouts. The numeric- and verbal-receipt rows fit on announced histories and read unannounced histories with the post-event receipt encoding held fixed; the verbal-to-numeric rows fit on the verbal receipt and read the numeric one. Recorded actions and first-post-event geometry are matched. Retained excess is the transferred AUROC excess over chance divided by the same-condition excess.}
\label{tab:receipts-history}
\end{table*}

\begin{table*}[t]
\centering
\footnotesize
\setlength{\tabcolsep}{3pt}
\renewcommand{\arraystretch}{1.12}
\begin{tabular}{llccc}
\toprule
\textbf{Checkpoint} & \textbf{Comparison} & \textbf{Same AUROC} & \textbf{Transfer AUROC} & \textbf{Retained excess} \\
\midrule
Qwen3-4B & Numeric receipt & 0.670 & 0.695 & 1.146 \\
 & Verbal receipt & 0.758 & 0.798 & 1.154 \\
 & Numeric $\to$ verbal & 0.670 & 0.661 & 0.947 \\
 & Verbal $\to$ numeric & 0.758 & 0.693 & 0.747 \\
\addlinespace[3pt]
Qwen3-8B & Numeric receipt & 0.692 & 0.672 & 0.900 \\
 & Verbal receipt & 0.729 & 0.709 & 0.912 \\
 & Numeric $\to$ verbal & 0.692 & 0.649 & 0.776 \\
 & Verbal $\to$ numeric & 0.729 & 0.682 & 0.798 \\
\addlinespace[3pt]
Gemma-3-4B & Numeric receipt & 0.579 & 0.532 & 0.406 \\
 & Verbal receipt & 0.622 & 0.548 & 0.392 \\
 & Numeric $\to$ verbal & 0.579 & 0.555 & 0.699 \\
 & Verbal $\to$ numeric & 0.622 & 0.505 & 0.038 \\
\addlinespace[3pt]
GPT-oss-20B & Numeric receipt & 0.942 & 0.960 & 1.040 \\
 & Verbal receipt & 0.951 & 0.924 & 0.940 \\
 & Numeric $\to$ verbal & 0.942 & 0.931 & 0.974 \\
 & Verbal $\to$ numeric & 0.951 & 0.872 & 0.824 \\
\addlinespace[3pt]
Qwen3-4B + maze RL & Numeric receipt & 0.693 & 0.665 & 0.852 \\
 & Verbal receipt & 0.798 & 0.778 & 0.931 \\
 & Numeric $\to$ verbal & 0.693 & 0.643 & 0.739 \\
 & Verbal $\to$ numeric & 0.798 & 0.652 & 0.510 \\
\bottomrule
\end{tabular}
\caption{Coherent background symbol-rule changes at each checkpoint's fixed site, with the numeric or verbal receipt still written out. Every row changes the background symbol rule between the fit and the read: the first two keep the receipt encoding, the last two change it as well. Columns and layouts are as in Table~\ref{tab:receipts-history}.}
\label{tab:receipts-background}
\end{table*}

\paragraph{Secondary comparisons.}
Table~\ref{tab:receipts-history} reports the history and reverse-transfer comparisons and Table~\ref{tab:receipts-background} the background-mapping comparisons, in which the symbol-to-score rule and board rendering are coherently changed while the receipt stays explicit. GPT-oss-20B's incremental reference-target share is $0.011$ on both trajectory sets.

\end{document}